\documentclass[a4paper,twoside]{article}

\usepackage{epsfig}
\usepackage{subcaption}
\usepackage{calc}
\usepackage{amssymb}
\usepackage{amstext}
\usepackage{amsmath}
\usepackage{amsthm}
\usepackage{multicol}
\usepackage{multirow}
\usepackage{pslatex}
\usepackage{subcaption}
\usepackage{apalike}
\usepackage{hyperref}
\usepackage[bottom]{footmisc}

\usepackage{array, boldline, makecell, booktabs}

\usepackage{SCITEPRESS}     

\usepackage[table]{xcolor}
\definecolor{lightgray}{gray}{0.93}

\begin{document}

\title{WSAM: Visual Explanations from Style Augmentation as Adversarial Attacker and Their Influence in Image Classification}

\author{\authorname{Felipe Moreno-Vera*\sup{1}\orcidAuthor{0000-0002-2477-9624}, Edgar Medina*\sup{2}, and Jorge Poco\sup{1}\orcidAuthor{0000-0001-9096-6287}}
\affiliation{\sup{1}Fundação Getúlio Vargas, Rio de Janeiro, Brazil}
\affiliation{\sup{2}QualityMinds, Munich, Deutschland}
\affiliation{*~means equal contribution}
\email{\{felipe.moreno, jorge.poco\}@fgv.br, edgar.medina@qualityminds.de}
}

\keywords{Style augmentation, adversarial attack, understanding, style, convolutional networks, explanation, interpretability, domain adaptation, image classification, model explanation, model interpretation.}

\abstract{Currently, style augmentation is capturing attention due to convolutional neural networks (CNN) being strongly biased toward recognizing textures rather than shapes. Most existing styling methods either perform a low-fidelity style transfer or a weak style representation in the embedding vector. This paper outlines a style augmentation algorithm using stochastic-based sampling with noise addition to improving randomization on a general linear transformation for style transfer. With our augmentation strategy, all models not only present incredible robustness against image stylizing but also outperform all previous methods and surpass the state-of-the-art performance for the STL-10 dataset. In addition, we present an analysis of the model interpretations under different style variations. At the same time, we compare comprehensive experiments demonstrating the performance when applied to deep neural architectures in training settings.}

\onecolumn \maketitle \normalsize \setcounter{footnote}{0} \vfill

\section{\uppercase{Introduction}}
\label{sec:introduction}

Currently, deep learning neural nets require a large amount of data, usually annotated, to increase the generalization and obtain high performance. To deal with this problem, methods for artificial data generation are performed to increase the training samples; this common learning strategy is called data augmentation. In computer vision, data augmentation increases the number of images through pixel-level processing and transformations. For supervised tasks where labels are known, these operations perform label-preserving transformations controlled by the probability of applying the operation and usually a magnitude that intensifies the operation effects on the image~\cite{Szegedy2016,Tanaka2019}. More recently, random erasing~\cite{DeVries2017a} and GAN-based augmentation~\cite{Tanaka2019} improved the previous accuracy. In contrast, recent advances in style transfer~\cite{Ghiasi2017,Jackson2018} lead us to think about the influence of applying random styling and what deep networks learn from this.

Style augmentation is a technique that generates variations from an original set of images changing only the style information and keeping the main content. The style transformation applied to the image changes the image's pixel information, generating a new diverse set of samples that follow the same original distribution. In contrast, content information remains equal~\cite{Ghiasi2017}. However, original style transfer techniques started with heavy computation to generate one stylized image. Experimentally, augmenting the training set randomly shows a new level of stochastic behavior, avoids overfitting in a small dataset, and stabilizes performance on large ones~\cite{Zheng2019}. Nowadays, some can work close to real-time performance while others can generate a batch of styles per image~\cite{Ghiasi2017,Jackson2018}. 


In Interpretable Machine Learning (IML), specifically in image-based models such as CNN, several methods exist to interpret and explain predictions. Usually, large and complex models like CNN are called ``black-box" due to their vast number of parameters (hidden layers). So, to know the information shared through each layer, some methods were developed using information from layers and gradients such as Saliency Maps~\cite{simonyan2013deep}, and CAM-based methods~\cite{zhou2016learning,selvaraju2017grad}. These methods help explain complex ``black box" image-based models and identify essential features in each sample prediction. In our approach, we will use these model explainers to highlight regions inside the input images to provide a visual interpretation of them.

In this work, we propose an augmentation strategy based on traditional augmentation plus style transformations. Besides, we implement new methods to visualize, explain, and interpret the behavior of our trained models. Also, we can understand which features are activating based on the style augmentation selected and study the influence of that style. Our main contributions in the present work are summarized as follows:

\begin{itemize} 
    \item We give an explanation of the successful augmentation strategy based on interpretation methods.
    \item We propose a \textbf{Style Activation Map} (SAM), \textbf{Weighted Style Activation Map} (WSAM), and \textbf{WSAM Variance} to visualize and understand the influence of style augmentation. 
    \item We outperform previous results on the STL-10 dataset using traditional and style augmentations.
\end{itemize}


\section{\uppercase{Related Works}}

\label{sec:relatedworks}
\subsection{Style Transfer}
\label{ssec:styletransfer}

In the first neural algorithm~\cite{Gatys2015}, a content image and a style image are inputted to the neural network to obtain an output image with the original content but a new style.  \cite{Jing2017} employed the Gram matrices to model textures by encoding the correlations between convolutional features from different layers. Previous style transfer works \cite{Ulyanov2017} improve the visual fidelity in which semantic structure was preserved for images with higher resolution. In~\cite{Geirhos2018} was concluded that neural networks have a strong bias with texture. Although the initial developments generated exciting results compared to the pioneer method, drawbacks such as weak texture synthesis and high computational cost were present \cite{Ulyanov2017,Jing2017}. More recently, \cite{Li2018,Ghiasi2017} solved the problem by relying on arbitrary styles without retraining the neural model. Also, other techniques adjusted a new parameter or inserted noise carefully to generate more style variations from one style input \cite{Ghiasi2017,kotovenko2019iccv}. Using these latter strategies, the first augmentation employing successfully style augmentation performing a cross-domain classification task \cite{Jackson2018} follows the methodology adopted on \cite{Ghiasi2017}, which uses an Inception-v3 \cite{Szegedy2015} architecture for the encoder and residual blocks for the decoder networks. However, the latent space is modified by a multivariate normal distribution which changes the style embedding. Other contemporary approaches \cite{Zheng2019,Georgievski2019} used style augmentation and reported exciting results in classification tasks, specifically STL-10, CIFAR-100, and Tiny-ImageNet-200 datasets. Other interesting applications are extended to segmentation tasks \cite{Hesse2019a,Gkitsas2019}.

Based on this literature review, we used a neural transfer model following a trade-off between edge preservation, flexibility to generate style variations, time processing, and best visual fidelity under different styles. We also compare our methodology to prior approaches used for style augmentation.

\subsection{Deep Network Explanations}

Explaining a CNN is focused on analyzing the information passed through each layer inside the network. Following this idea, several methods were proposed to visualize and obtain a notion about which features of a deep CNN were activated in one specific layer. In~\cite{simonyan2013deep} (saliency maps) showed the convolutional activations, \cite{zeiler2014visualizing} showed the impact of applying occlusion to the input image. In other methods, they use the gradients to visualize features and explain deep CNN networks such as DeepLIFT \cite{shrikumar2017learning}, which computes scores for each feature; Integrated Gradients \cite{sundararajan2017axiomatic}, which computes features based on gradients; CAM \cite{zhou2016learning}, and Grad-CAM~\cite{selvaraju2017grad} which computes relevant regions using gradient and feature maps. Each method identifies features with high and strong activation representing the prediction for a specific predicted category.

Guided by this literature review, we propose a new method called \textbf{Style Activation Maps (SAM)} based on the Grad-CAM method applied to style augmentation. We choose this one due to better behavior and performance against adversarial attacks or noise-adding techniques \cite{adebayo2018sanity,gilpin2018explaining}. Our main goal is to understand and interpret the impact of applying style augmentation in classification tasks and analyze their influence. 

\section{\uppercase{Proposed Method}}

In this section, we present theoretical formulation and some interpretation methods used.

\subsection{Style Augmentation}

For our experiments, we used the same methodology as \cite{Jackson2018}; we nevertheless used a faster VGG-based network and added noise to diversify the style features. Specifically, we used an architecture composed of a generalized form of a linear transformation \cite{Li2018}. Also, we compare with other related works \cite{Jackson2018,Zheng2019} that use neural style augmentation.

Formally, let $C=\big\{ c_1, c_2, ..., c_j\big\}, c_i \in \mathbb{R}^{N \times M \times C}$ be the content image set and let $Z=\big\{z_1, z_2, ..., z_i \big\}, z_i \in \mathbb{R}^{n}$ be the precomputed style embedding set from $S=\big\{s_1, s_2, ..., s_i \big\}, s_i \in \mathbb{R}^{N \times M \times C}$, are used to feed the styling algorithm to generate the output set $O=\big\{ o_1, o_2, ..., o_j\big\}, o_j \in \mathbb{R}^{N \times M \times C}$. Moreover, we denote zero-mean vectors $\overline{c}_j \in \mathbb{R}^{N \times M \times C}$ and $\overline{z}_i \in \mathbb{R}^{n}$. Our style strategy transfers elements $z_i$ from the style set $Z$ to a specific element from the content set $C$.

The VGG (``r41") architecture, denoted as $M(.)$, maps $\mathbb{R}^{N \times M \times C} \rightarrow \mathbb{R}^{N_1 \times M_1 \times F}$. and a non-linear function $\phi(.)$ maps $\mathbb{R}^{N_1 \times M_1 \times F_1} \rightarrow \mathbb{R}^{n}$, where $N_1<N$, $M_1<M$ and $F_1>F$. Also, we denote $C(.)$, $U(.)$ as the compress and uncompress CNN-based networks from the original paper \cite{Li2018}. $\phi(.)$ embeds the input image to an embedding vector that contains the semantic information of the image. More concisely, we use this non-linear function to map the original image to an embedding vector as shown in Eq.~\ref{embedding1} for the content image and Eq. \ref{embedding2} for the style image. In our implementation, the function $\phi(.)$ employs a CNN whose output is used to compute the covariance matrix and feed it to a fully-connected layer.

Since we use an architecture based on linear transformations, which is generalized from previous approaches~\cite{Ghiasi2017}, the transformation matrix $T$ sets and preserves the feature affinity of the content image (determined by the covariance matrix of the content and the style). This is expressed in Eq.~\ref{transform}. In our implementation, we precomputed the style vectors and saved all textures in memory; thereby, our modifications are described in Eq.~ \ref{implementation}~and~\ref{return}.

\begin{gather}
\phi_c = \phi_1(VGG(\overline{c}_j)) \label{embedding1}\\
\phi_s = \phi_2(VGG(\overline{s}_i)) \label{embedding2} \\
T = \phi_c\phi_c^T   \phi_s\phi_s^T \label{transform}
\end{gather}

In our implementation, we precompute the style vector and save all textures in memory; thereby, our modifications are expressed in Eq.~\ref{implementation}~and~\ref{return}. 

\begin{equation}
    T = \phi_c\phi_c^T  ( \alpha \phi_c\phi_c^T + (1-\alpha) \hat{z_i}) 
    \label{implementation}
\end{equation}
\begin{equation}
    o_i = U(T \hspace{0.1cm} C(c_j)) + (\alpha)\mu_{c_i} + (1-\alpha)\mu_{z_i}
    \label{return}
\end{equation}

Where $\alpha$ is the interpolation hyper-parameter which controls the strength of the style transfer similarly to \cite{Jackson2018}, and $\hat{z_i}$, defined in Eq.~\ref{gaussian}, is the embedding vector of the style set with a noise addition for style randomization.

\vspace{-0.25cm}
\begin{equation}
    \hat{z_i} \sim \overline{z}_i + \mathcal{N}(\mu_i,\,\sigma_i^{2})
    \label{gaussian}
\end{equation}

As argued in prior methodologies, minor variations increase the randomization in the process; thereby, we apply noise instead of using a sampling strategy similar to applying a Gaussian noise in the latent space of generative networks during the training. In particular, we set this noise source as a multivariate normal distribution which means covariance scales and shifts $\overline{z}_i$ into the embedding space. This is also useful for understanding the randomization process and the influence of the latent space.


\subsection{Model Interpretation}



In this work, we propose a new method \textbf{Style Activation Map} based on Grad-CAM to visualize the predictions and the highlighting regions with the most representative activated features from styled images. To do this, we extract from the penultimate layer the $A^k \in \mathbb{R}^{u \times v}$ feature maps of width \textit{u} and height \textit{v}, with each element indexed by \textit{i},\textit{j}. So $A^k_{i,j}$ refers to the activation at location $(i,j)$ of the feature map $A^k$. We apply the GlobalAveragePooling (GAP) technique to the feature maps to get the neuron importance weights defined in Eq.~\ref{feature_maps_eq}.

\begin{equation}
    \delta^c_k = \overbrace{ \frac{1}{Z} \sum_i \sum _j}^{\text{GAP}} \underbrace{\frac{\partial y^c}{\partial A^k _{ij}}}_\text{grad-backprop}
    \label{feature_maps_eq}
\end{equation}

Where $\delta^c _k $ represents the neuron importance weights, \textit{c} is the class, $Z=u\times v$ the size of the image, \textit{k} is the k-th feature map, $A^k_{ij}$ is the feature map, $y^c$ the score for class \textit{c}, and $\frac{\partial y^c}{\partial A^k _{ij}}$ is the gradient vector obtained via back-propagation. Next, we calculate the corresponding activation maps for each prediction using Eq.~\ref{feature_maps_eq}. From this point, we propose a new technique to visualize the highlighted regions in stylization and their variations. We present two methods: the \textbf{Style Activation Map (SAM)} defined as the relevant highlighted regions of the different styles in predictions and the \textbf{Weighted Style Activation Map (WSAM)} defined as the weighted sum of all styles applied in all samples per class.

We denote the \textbf{SAM} of a styled image with style $\sigma$, style intensity $\alpha$, and class \textit{c} by $I_{\alpha, \sigma}^c$. Where $\alpha$ is the style intensity used, $\sigma$ is the style used. Also, we have the \textit{k}-th feature activation maps $A^k \in \mathbb{R}^{u \times v}$, and their class score $y^c$ for class \textit{c}:

\begin{equation}
    SAM_{\alpha, \sigma}^c = ReLU(\sum_k \delta^c _k A_{\alpha,\sigma}^k)
    \label{cam_eq}
\end{equation} 

We apply the ReLU function to the weighted linear combination of the feature maps $A^k$ because we are only interested in features with a \textbf{positive influence}. Then, we use this result to obtain the \textbf{WSAM} doing a weighted mean of $SAM^c_{\alpha,\sigma}$ and their predictions $y^c_{\alpha,\sigma}$ using all styles and all intensities. We define $\Omega$ as the product of total styles and total intensities evaluated, so we have:

\begin{equation}
    WSAM^c = \frac{1}{\Omega} \sum_\alpha \sum_\sigma y^c_{\alpha,\sigma} \times SAM_{\alpha, \sigma}^c
    \label{wsam_eq}
\end{equation}

Once we calculate $WSAM^c$ in eq. \ref{wsam_eq} we will calculate the total variance region of $m$ samples to identify the most significant styles features for the classifier:

\begin{equation}
    WSAM_{variance}^c = \frac{1}{Z \times m} \sum^m_i ( WSAM^c_i - y^c_i \times I^c_i)^2
    \label{wsam_mse_eq}
\end{equation}

Where $I^c_i$ is the i-th input sample stylized with $\alpha=1.0$ (no style), $Z = u \times v$ the image size, and their class score $y^c_i$ for the class \textit{c}. Our metric shows the highlighted region variance between an image and its styles with different $\alpha$s. 

\begin{figure}[h!]
    \centering
    \begin{subfigure}{0.4\textwidth}
            {\epsfig{file = 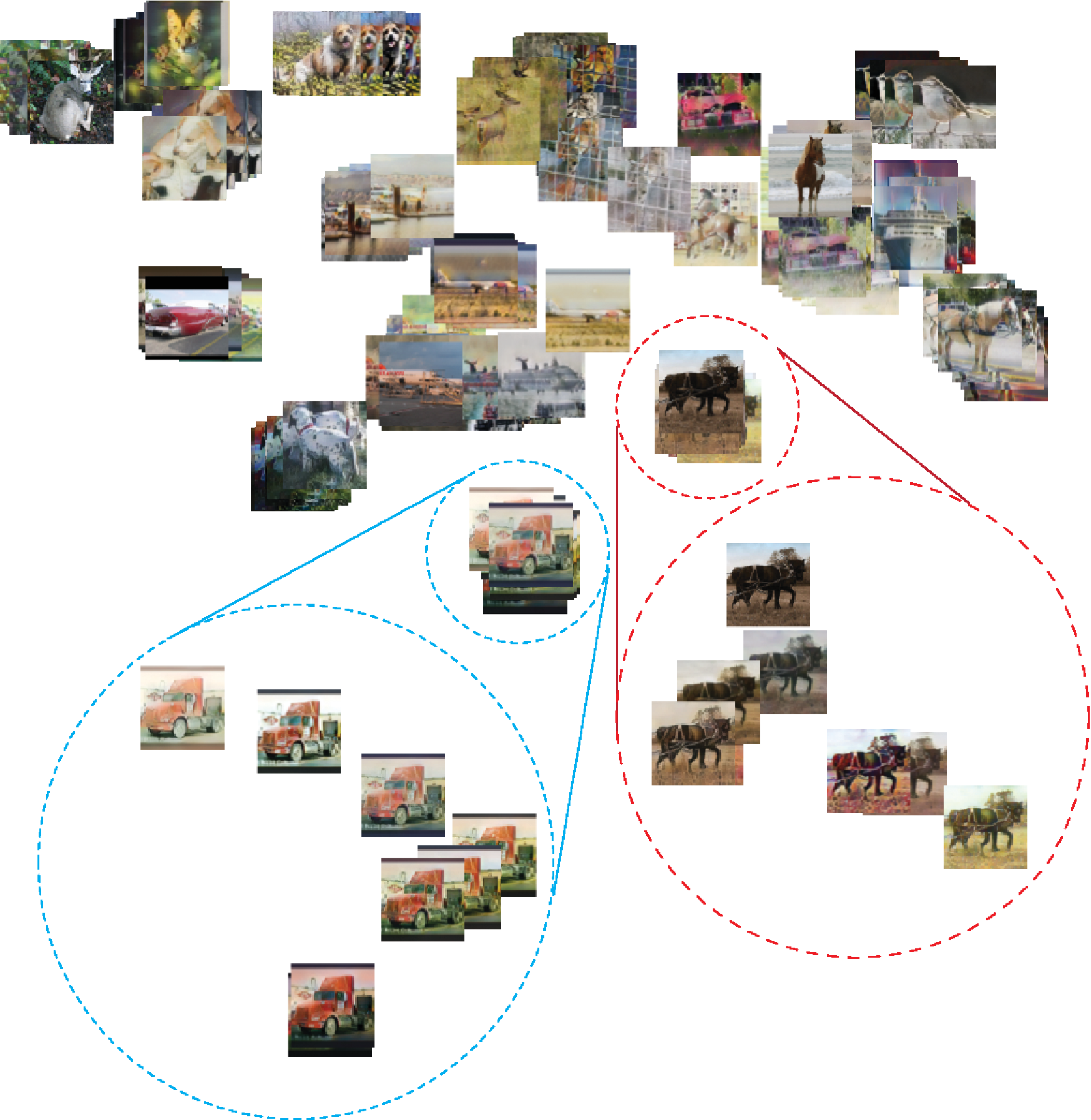, width = 6cm}}
    	\caption{}
    	\label{fig:tsne}
    \end{subfigure}
    \par\bigskip
    \begin{subfigure}{0.5\textwidth}
        \centering
    	{\epsfig{file = 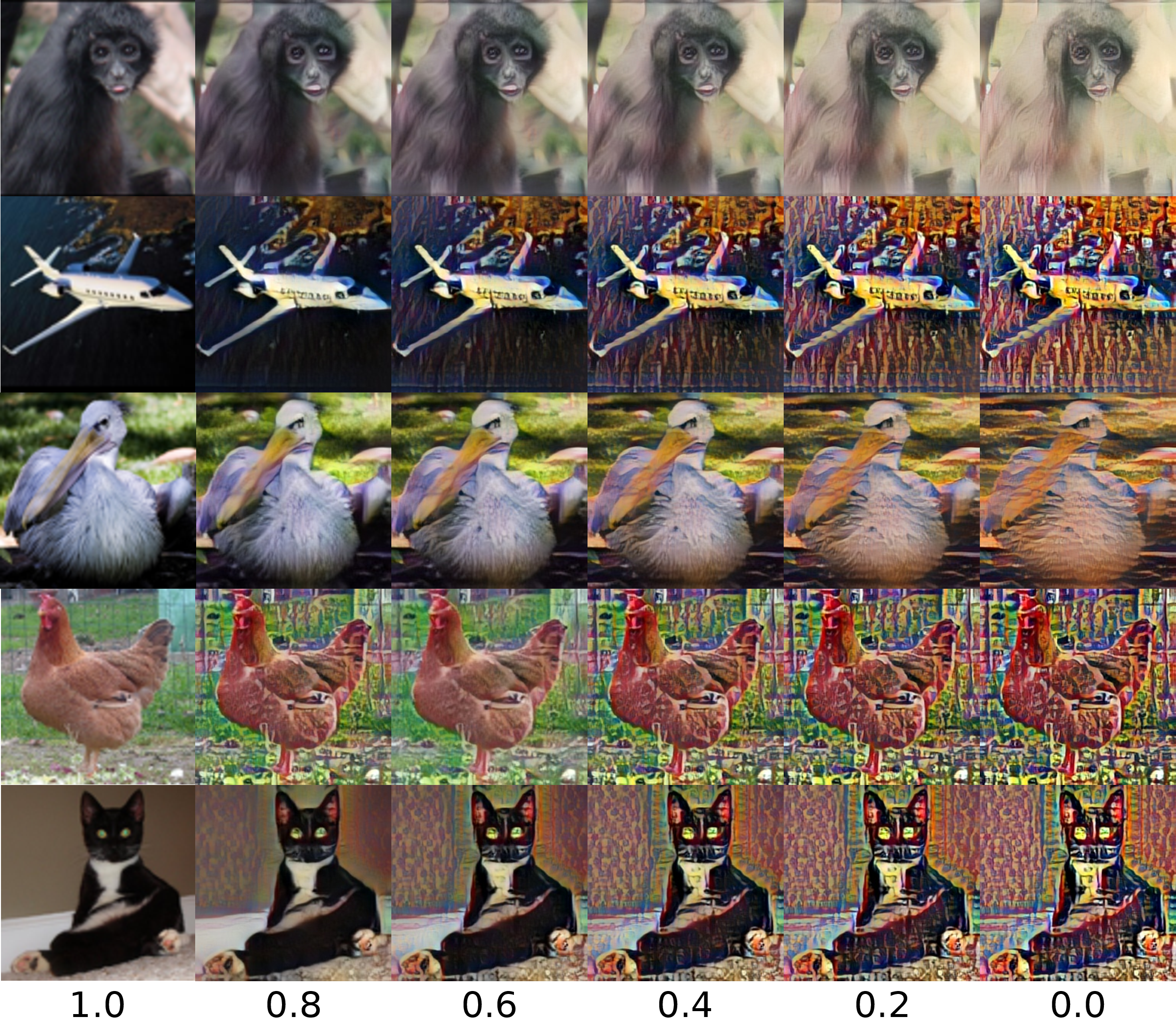, width = 6cm}}
    	\caption{}
    	\label{fig:style_aug}
    \end{subfigure}
	\caption{(a) Visualization using t-SNE, samples with their style augmentation. (b) Different styles with variations of the parameter $\alpha$ from 1.0 (no stylization) to 0.0 (style augmentation) on images.}
\end{figure}

\section{\uppercase{Experiments and Results}}
We perform our experiments using the STL-10 ($96 \times 96$) dataset, where samples are distributed in 5,000 and 8,000 labeled data for training and testing, respectively. We disregard the 100,000 unlabeled data for all our experiments. Besides, all experiments were performed using five different networks with high performance, such as Xception~\cite{Chollet2016}, InceptionV3-299~\cite{Szegedy2015}, InceptionV4~\cite{Szegedy2016}, WideResNet-96~\cite{Zagoruyko2016}, and WideResNet-101~\cite{spinal2022net}. We also compare our results with other state-of-the-art style augmentation like SWWAE~\cite{Zhao2015}, Exemplar Convnet~\cite{Dosovitskiy2014}, IIC~\cite{Ji2018}, Ensemble~\cite{Thoma2017}, WideResNet+cutout~\cite{DeVries2017a}, InceptionV3~\cite{Jackson2018}, and STADA~\cite{Zheng2019}. 

\subsection{Style Augmentation}

First, we explore the effects of style augmentation through t-SNE visualization of images after applying the styler network to a subset of the test set Figure~\ref{fig:tsne}; we note some clusters of original images and their styles separate a bit of distance such as truck and horse. In Figure~\ref{fig:style_aug}, we performed some styles using some $\alpha$ values to find the best balance between style and content information as described above in Eq.~\ref{return}. However, we emphasize the difference between traditional augmentation and the classical technique like rotation, mirroring, cutout~\cite{DeVries2017a}, etc. With style augmentation, we increase the number of samples using about 80 000 styles and sampling for style intensity. Figure~\ref{fig:style_aug} shows different styles and different $\alpha$ values (style intensity) from 0.0 to 1.0 by steps of 0.2. 


Images with augmentation strategies for training deep models include traditional augmentations, cutouts, and our style augmentation method using a lower style effect ($\alpha = 0.7$). At this point, we can consider \textbf{style augmentation} as a \textbf{noise-adding technique} or \textbf{adversarial attacker} due to the style distortion, which makes images more challenging to represent and be associated with the correct class.


\subsection{Training Models}
\label{ssec:experiments_trainedmodels}
For experiments, we define four learning strategies, which are composed of no augmentation (None or N/A), traditional augmentation (Trad), style augmentation (SA), and both (Trad+SA) for each model. In Table~\ref{AccComparison}, we present the quantitative comparisons between the state-of-the-art methods in style augmentation using styling and architectures for the STL-10 dataset; the Extra column means additional data is used to train that model, Trad column means traditional augmentation plus cutout, and Style column indicates our style augmentation.

\newcolumntype{g}{>{\columncolor{lightgray}}c}
\begin{table}[ht]
\small
\rowcolors{1}{}{white}
\begin{tabular}{lgggg}
\toprule
\rowcolor{white}
Network & Extra & Trad & Style & Acc \\
\midrule
SWWAE& \checkmark & \checkmark & & 74.33 \\
Exempla Conv & \checkmark & \checkmark & & 75.40 \\ 
IIC& \checkmark & \checkmark & & \textbf{88.80} \\
Baseline& & \checkmark & & 75.67 \\
Ensemble& & \checkmark & & 77.62 \\	
STADA$^*$& & \checkmark & \checkmark & 75.31 \\
InceptionV3-299$^{*}$& & \checkmark & \checkmark & \textbf{80.80} \\
Xception-96$^{*}$& & \checkmark & \checkmark & \textbf{82.67} \\
Xception-128$^{*}$& & \checkmark & \checkmark & \textbf{85.11} \\
\midrule
& & & & 73.37 \\
& & \checkmark & & 86.19 \\
& & & \checkmark & 74.89 \\
\multirow{-4}{*}{Xception-256*}& & \checkmark & \checkmark & \textbf{86.85} \\
\midrule
& & & & 79.17 \\
& & \checkmark & & 86.49 \\
& & & \checkmark & 80.52 \\
\multirow{-4}{*}{InceptionV4-299*}& & \checkmark & \checkmark & \textbf{88.18} \\
\midrule
& &  &  & 77.28 \\
& & \checkmark & & 87.26 \\
& & & \checkmark & 83.58 \\
\multirow{-4}{*}{\makecell[l]{WideResNet-96*\\(WRN)} }& & \checkmark & \checkmark & \textbf{88.83} \\
\midrule
& &  &  & 87.83 \\
& & \checkmark & & 88.23 \\
& & & \checkmark & 92.23 \\
\multirow{-4}{*}{\makecell[l]{WideResNet-101*\\(WRN)} }& & \checkmark & \checkmark & \textbf{94.67} \\
\bottomrule
	\end{tabular}
	\caption{Accuracy comparison of data augmentation methods in STL-10 ($^{*}$) indicates results performed by us.} 
	\label{AccComparison}
\end{table}

\begin{figure}[h!]
    \centering
    \begin{subfigure}{0.4\textwidth}
            {\epsfig{file = 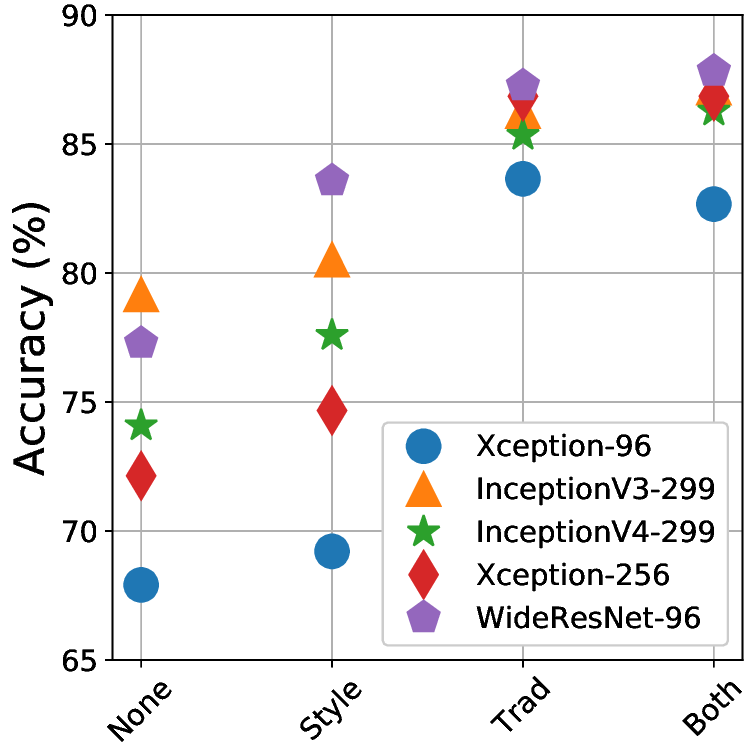, width = 6cm}}
    	\caption{}
    	\label{fig:comparisonAcc}
    \end{subfigure}
    \par\bigskip
    \begin{subfigure}{0.5\textwidth}
        \centering
    	{\epsfig{file = 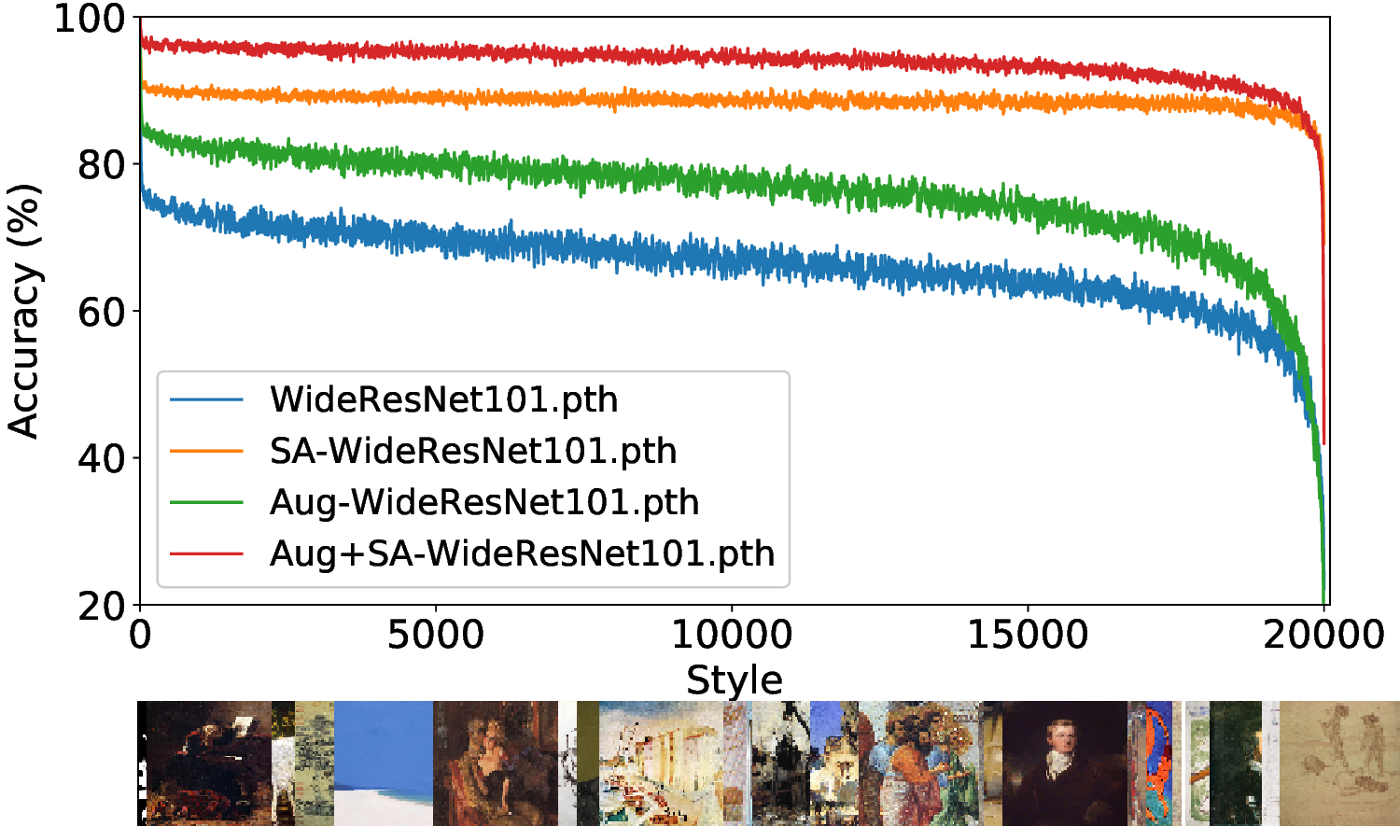, width = 8cm}}
    	\caption{}
    	\label{fig:StyleInfluence}
    \end{subfigure}
	\caption{(a) Influence of the application of styles on a subset of the test set. (b) Comparison of WideResNet-101 robustness under style augmentation setting during training. Accuracy vs. style transfer ($\alpha=0.5$) for a subset of the test set.} 
\end{figure}

We note that in all cases, style augmentation helps to improve results. Besides, we found that models with higher input resolution reached higher accuracy after applying the styling method shown in Figure \ref{fig:comparisonAcc}. Experiments on different input sizes support this affirmation \cite{Chollet2016}. 


Furthermore, in Figure \ref{fig:StyleInfluence}, we analyze the influence of style additions to a subset of the test set composed of 100 samples (10 samples per class), computing their average accuracy in each point on axis X consisting of an overall 20,000 random styles, these styles were sorted following from greater to lower accuracy. Note that the accuracy of the model trained without style augmentation decreased drastically for some styles. In contrast, the use of styles in training becomes the same architecture more robust to strong variations without losing accuracy.

\section{\uppercase{Style Activation Maps Visualization}}

\begin{figure}[h!]
    \centering
    \begin{subfigure}{0.4\textwidth}
            {\epsfig{file = 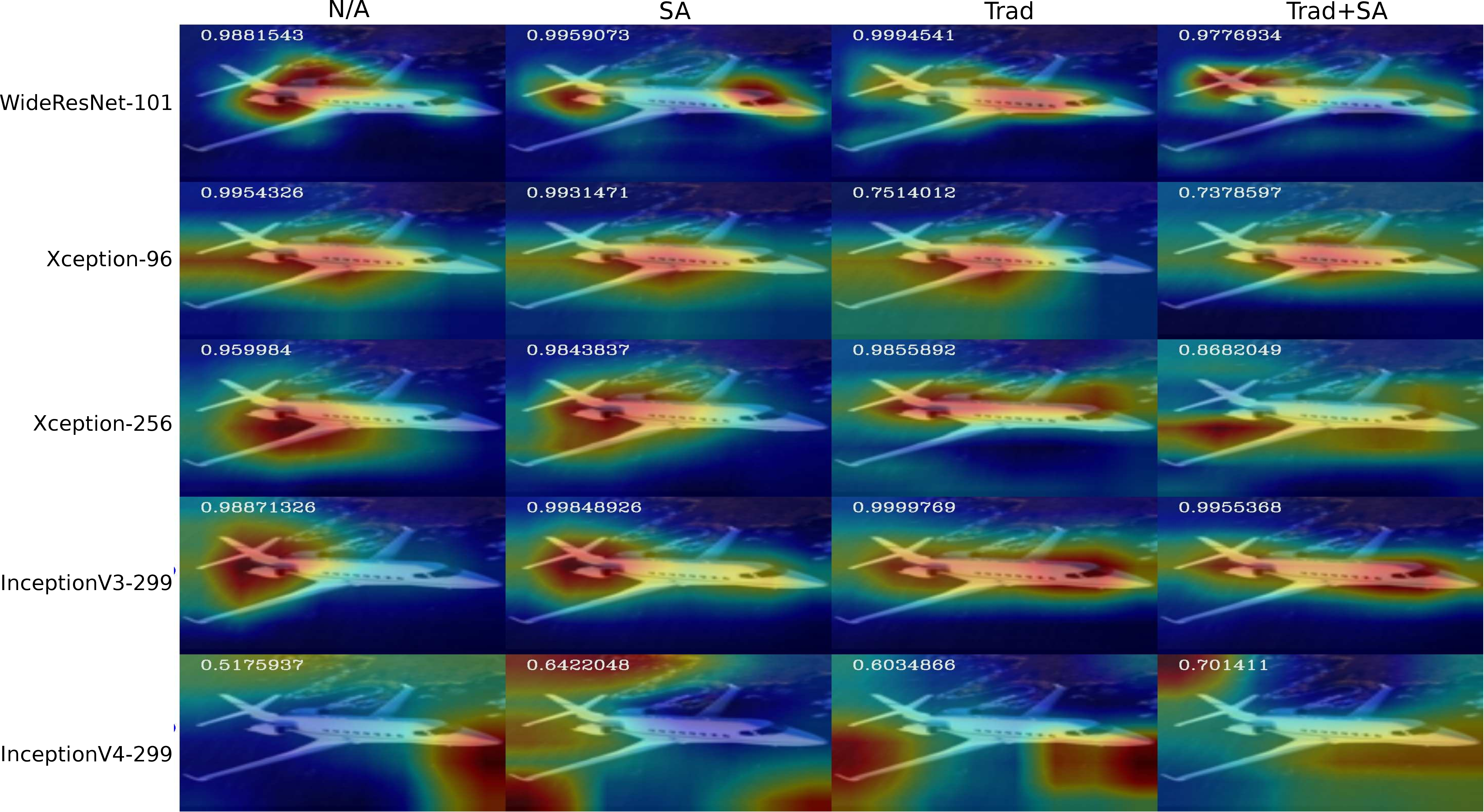, width = 6.5cm}}
    	\caption{}
    	\label{fig:models_cam_sample}
    \end{subfigure}
    \par\bigskip
    \begin{subfigure}{0.4\textwidth}
            {\epsfig{file = 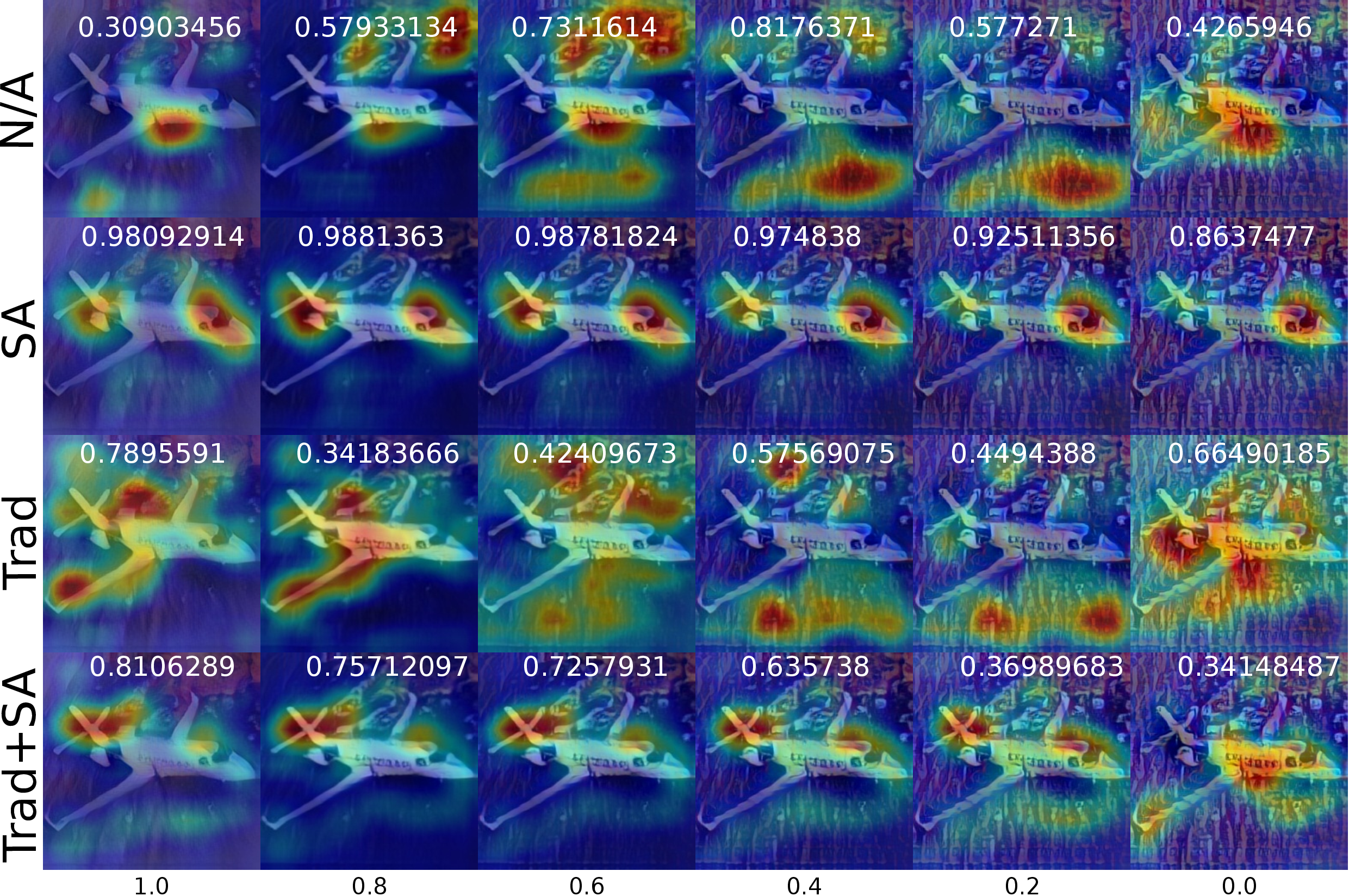, width = 6.5cm}}
    	\caption{}
    	\label{fig:style_cam_same_sample}
    \end{subfigure}
	\caption{Comparing SAM results: (a) We compare SAM from different models (rows) using the augmentations strategies: None, Trad, SA, and Trad+SA (columns). (b) We compare SAM of the WideResNet-101 trained using N/A, Trad, SA, and Trad+SA tested on the same image and style but varying the style intensity $\alpha$ as input.}
\end{figure}

Once our training step was finished, we evaluated and understood the stylization behavior in our models. First, in Figure~\ref{fig:models_cam_sample}, we show how our Style Activation Map works. Each row is the model, and each column is the learning strategy using no augmentation (N/A), using only style augmentation (SA), using traditional augmentation plus cutout (Trad), and using both (Trad+SA). We take a random sample with no style ($\alpha=1$) to calculate their SAM (style activation map) for each model and each augmentation strategy. From this, we see how both Trad and Trad+SA help the models to focus on the plane instead other regions like no augmentation (N/A). Also, it is important to highlight that the better the prediction, the more accurate the region in the object (in this case, a plane).

On the other hand, using the best model WideResNet-101, we use the same random sample (a plane) to test the different learning strategies using the same style but varying the $\alpha$ parameter. Let's say, in this case, we will use as input the stylized sample. In Figure~\ref{fig:style_cam_same_sample}, we show the influence of image stylization. Each row means learning strategy N/A, SA, Trad, and Trad+SA. So, each column implies that the styled input sample by $\alpha$ value varies from 1.0 (no intensity/style) to 0.0 (more intensity) before being evaluated by each network. We saw that a styled image tested in a model which does not use SA gets too bad results, but this did not happen in the model trained with SA. Also, the SAM-relevant regions in styled models tend to be constant along the $\alpha$ variations.

\begin{figure}[h!]
    \centering
    \begin{subfigure}{0.4\textwidth}
            {\epsfig{file = 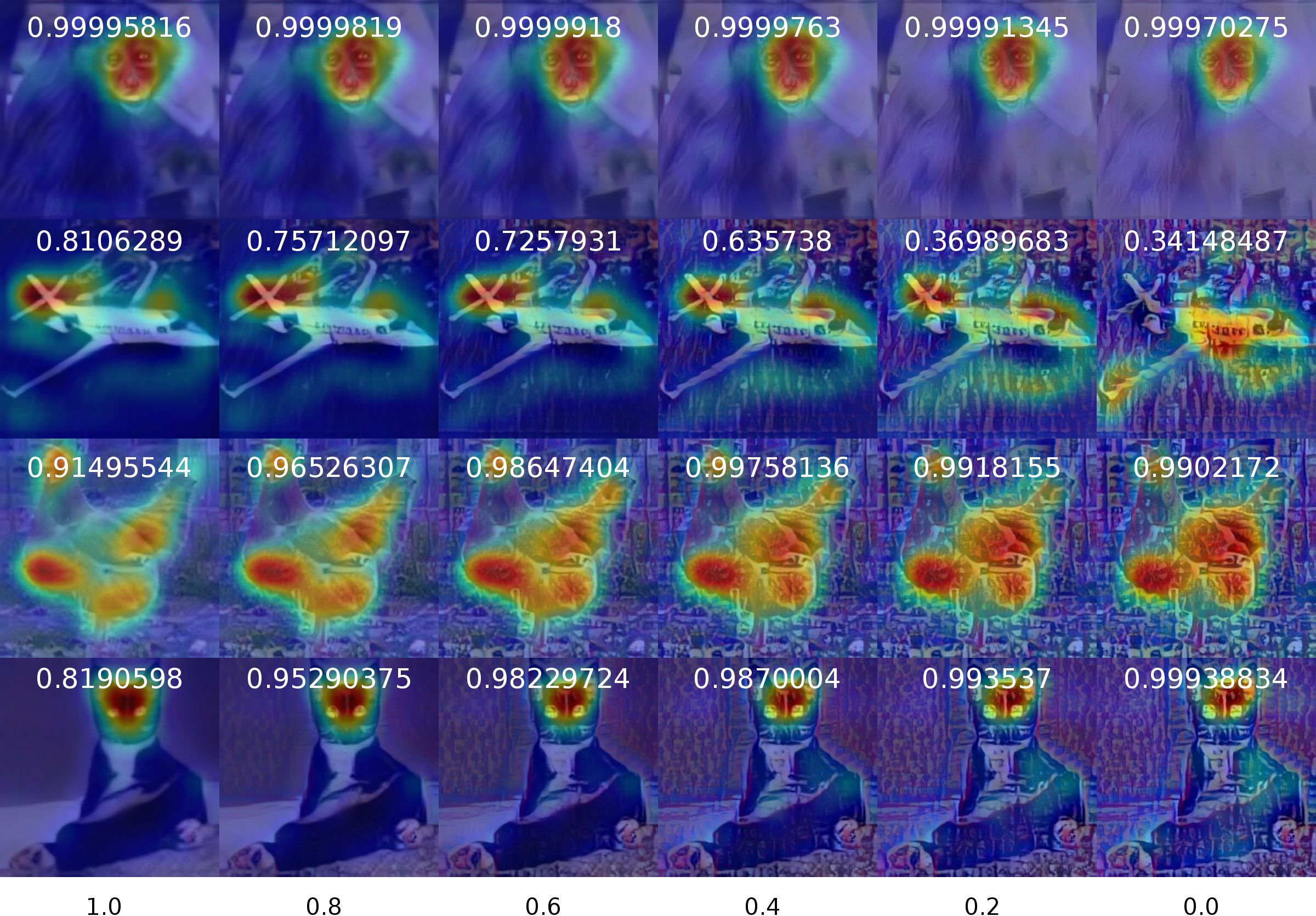, width = 6.5cm}}
    	\caption{}
	\label{fig:style_cam_different_samples}
    \end{subfigure}
    \par\bigskip
    \begin{subfigure}{0.4\textwidth}
            {\epsfig{file = 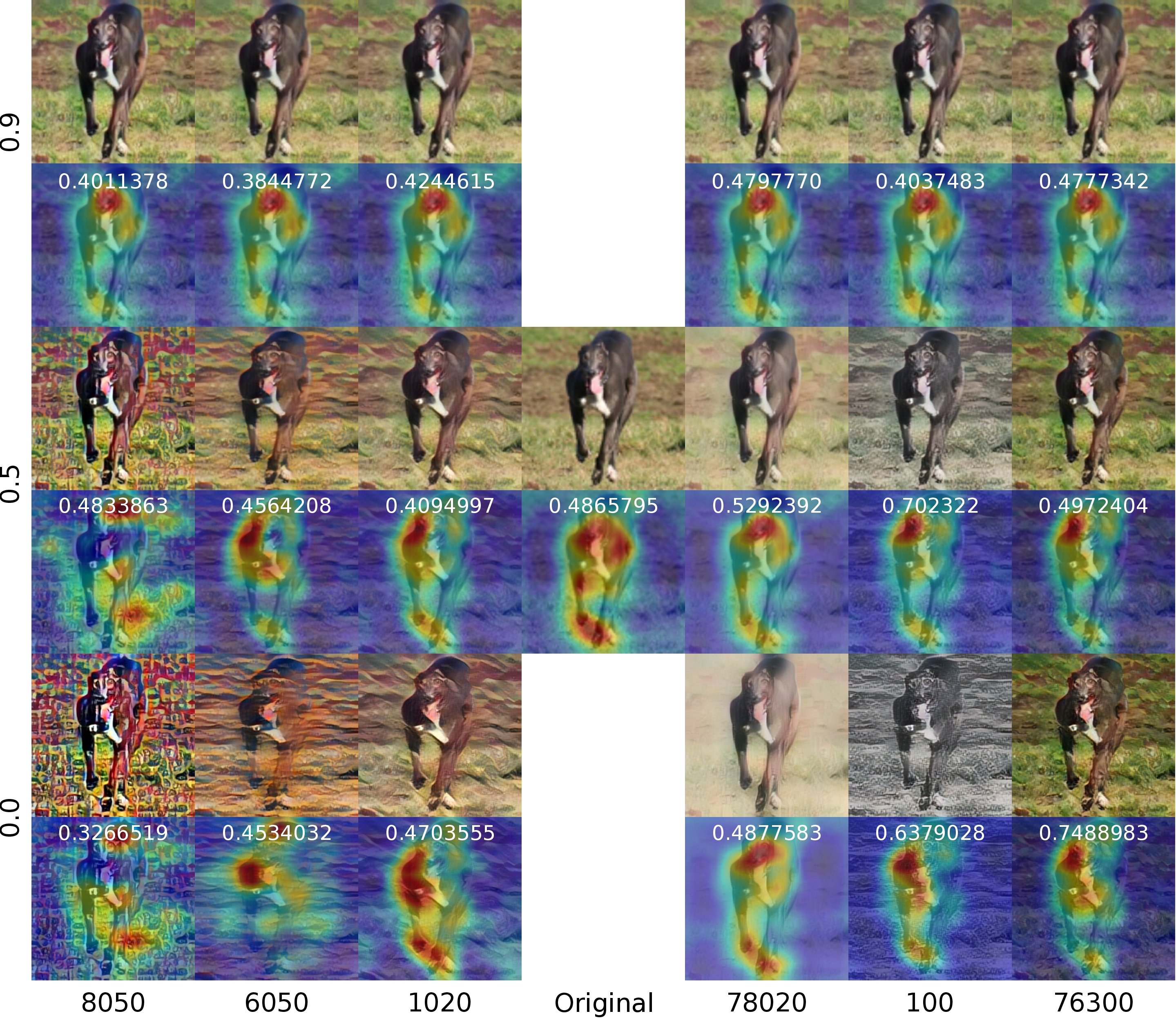, width = 6.5cm}}
    	\caption{}
    	\label{fig:styles_different_same_sample}
    \end{subfigure}
	\caption{Comparing SAM results: (a) WideResNet-101 SAM results using different values for $\alpha$, different styles, and different samples. (b) WideResNet-101 SAM results show the negative impact styles (3 on the left side), positive impact styles (3 on the right side), and no style evaluated for $\alpha=(0, 0.5, 0.9)$ to one input image (middle).}
\end{figure}

In Figure~\ref{fig:style_cam_different_samples}, we show different samples, styles, and $\alpha$ values. We show the influence of style in random samples with random styles; we note that some styles exist which don't help to improve prediction. Otherwise, it gets worse. From this result, we say some \textbf{styles can influence positively, negatively, or do not impact} the input image. In addition, this result shows how the relevant regions for the network change depending on the style, and these two results are shown in Figure~\ref{fig:styles_different_same_sample}, also improving or not the confidence of the prediction. 

\section{\uppercase{Discussions}}




%

We train, test, and visualize the impact of the style augmentation varying both $\alpha$ values (from 0.0 to 1.0 in steps of 0.2) and learning strategies (N/A, SA, Trad, and Trad+SA) using the STL-10 dataset. We achieve high performance and the best result with WideResNet-101. We show the behavior of the style augmentation technique proposed (see Figure~\ref{fig:tsne} and Figure~\ref{fig:style_aug}). We identify that some styles perturb the images more than others using the same sample, like adding noise. Also, we argue that by using larger input sizes and removing some complex styles, we probably remove the negative impacts on training (see Figure~\ref{fig:comparisonAcc}.
Furthermore, our experiments showed interesting robustness to styles when styling is included in the training (see Figure~\ref{fig:StyleInfluence}). Nonetheless, we also observed that the accuracy of models with Trad decreased drastically for some styles. Additionally, we found that some textures are more challenging to perform a style transfer using cutting-edge networks. 


We explored more deeply the effects of particular styles and their influence on training and testing. In Figure~\ref{fig:models_cam_sample}, we present how style made a model more robust thanks to the different intensities of $\alpha$, which behaves as noise but does not apply to everyone. Specifically, we took the case of the plane evaluated in Figure~\ref{fig:style_cam_same_sample}. We got a low accuracy (0.341\%) with higher style intensity ($\alpha = 0.0$). Otherwise, we got the highest accuracy (0.988\%) with  $\alpha = 0.8$. Furthermore, experiment results suggested that the best fit for $\alpha$ could be \textbf{between} 0.3 and 0.8, similar results were found in~\cite{Jackson2018}. In Figure~\ref{fig:style_cam_different_samples}, we note that some styles have no effects, and for others, the network learns how to classify images correctly with higher intensities (noise). Also, style strengthens the correlation between predictions and styled features activation maps (see Figure~\ref{fig:styles_different_same_sample}).

\begin{figure}[t!]
    \centering
    {\epsfig{file = 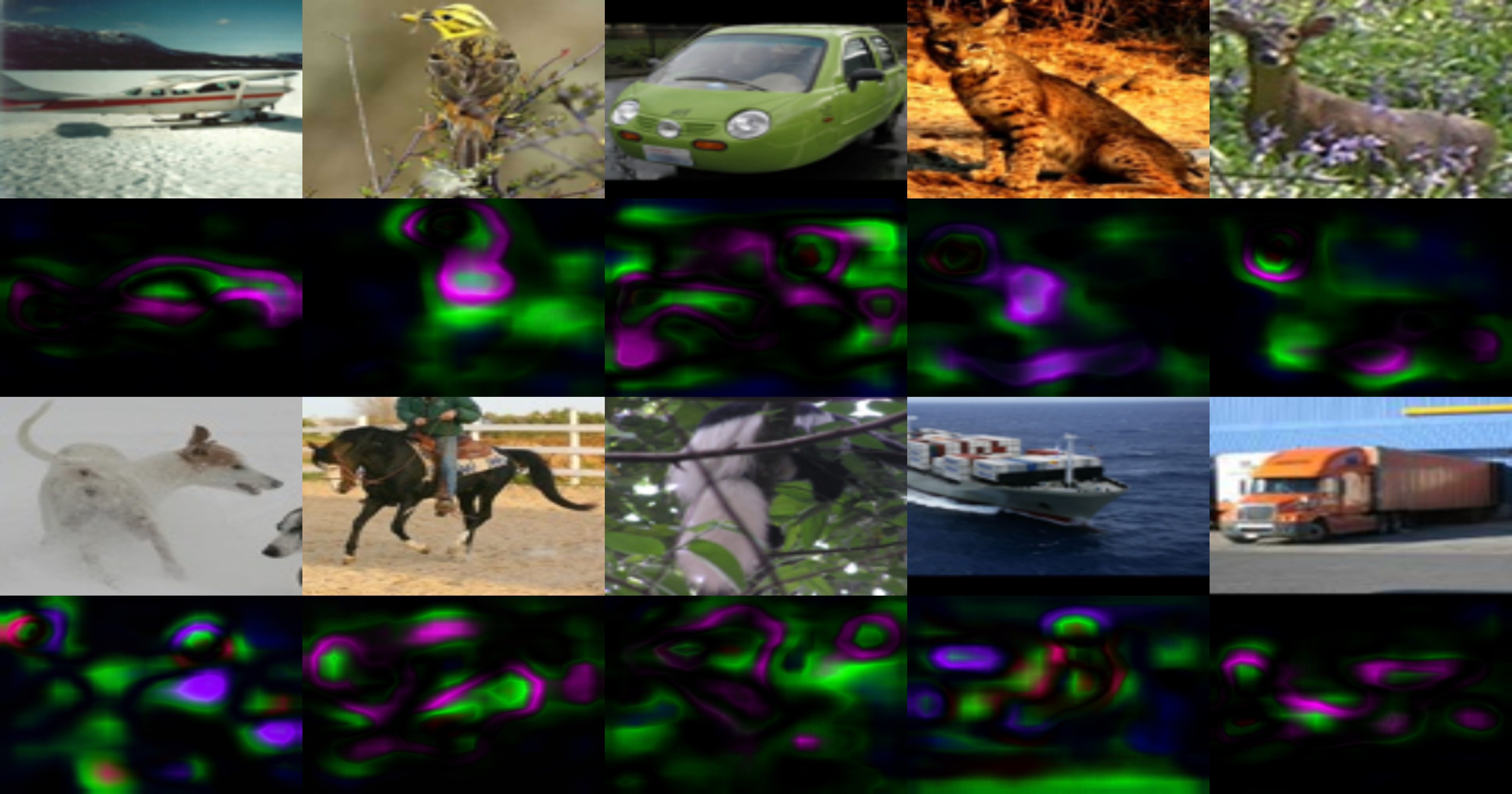, width = 7cm}}
    \caption{Results after calculating the WSAM for each class sample, varying styles, and $\alpha$ as defined in Eq.~\ref{wsam_eq}. We can see the total variance of the relevant region after stylization.}
    \label{fig:wsam_class}
\end{figure}

\begin{table}[h!]
\small
\centering
\rowcolors{1}{}{lightgray}
\begin{tabular}{cccc}
\toprule
$WSAM_{variance}$ & Category & $WSAM_{variance}$ & Category \\
\cmidrule(lr){1-2}\cmidrule(lr){3-4}
airplane & 0.107 & horse & 0.269 \\
truck & 0.129 & bird & 0.316 \\
deer & 0.175 & dog & 0.338 \\
cat & 0.193 & monkey & 0.380 \\
car & 0.228 & ship & 0.456  \\
\bottomrule
\end{tabular}
\caption{Results of the total WSAM variance sorted for each class in STL-10 after normalization.} 
\label{table:wsam_variance}
\end{table}

We now calculate the \textbf{WSAM variance} and WSAM for each class sample, using all styles and $\alpha$s. In Table~\ref{table:wsam_variance}, we present the \textbf{WSAM variance} of all \textbf{SAM}s. Besides, in Figure~\ref{fig:wsam_class}, we show the result of \textbf{WSAM} for one sample per class. These results give us an idea about the impact of applying \textbf{79 424} styles with different $\alpha$ intensities during the training phase and how the network learns to deal with those noisy samples (styled images), helping the robustness of the model. Finally, These results allow us to understand the influence of style augmentation in image classification. We can say that style augmentation can be used as a noise adder or adversarial attacker, making our model more robust against adversarial attacks.

\section{\uppercase{Conclusions and Future Work}}
\label{sec:conclusions}

In this work, we define a metric to explain by experimentation the behavior, the impact, and how the style augmentation may impact getting better results in the classification tasks. This metric is composed of three main outputs: Style Activation Map (SAM), Weighted Style Activation Map (WSAM), and $WSAM_{variance}$; this last one measures the variance of the regions of relevant features in styled samples. We outperform the state-of-the-art \textbf{without extra data} in style augmentation accuracy with WideResNet-101 trained on the STL-10 dataset; besides, our method gives robustness to input variations. From results and experiments, style augmentation has an impact on the model, and this impact can be visualized through SAM regions generated. We conclude that styles may modify and perturb different features from the input images (as an adversarial attacker), thus causing another set of images with slight variations in the distribution or becoming outliers making that prediction fail. In future directions, we will extend this study to more complex models with a higher number of parameters (like transformers) and higher images size like ImageNet and explain how style could influence their internal behavior. Also, we propose to understand more deeply which features are selected to be preserved in each style and which distortion they could generate through the network layers.

\section{\uppercase{acknowledgements}}
This work was supported by Carlos Chagas Filho Foundation for Research Support of Rio de Janeiro State (FAPERJ)-Brazil (grant \#E-26/201.424/2021), S\~ao Paulo Research Foundation (FAPESP)-Brazil (grant \#2021/07012-0), and the School of Applied Mathematics at Fundação Getulio Vargas (FGV/EMAp). 
Any opinions, findings, conclusions, or recommendations expressed in this material are those of the authors and do not necessarily reflect the views of the FAPESP, FAPERJ, or FGV.






\bibliographystyle{apalike}
{\small

\begin{thebibliography}{}

\bibitem[Adebayo et~al., 2018]{adebayo2018sanity}
Adebayo, J., Gilmer, J., Muelly, M., Goodfellow, I., Hardt, M., and Kim, B.
  (2018).
\newblock Sanity checks for saliency maps.

\bibitem[Chollet, 2016]{Chollet2016}
Chollet, F. (2016).
\newblock {Xception: Deep Learning with Depthwise Separable Convolutions}.

\bibitem[DeVries and Taylor, 2017]{DeVries2017a}
DeVries, T. and Taylor, G.~W. (2017).
\newblock {Improved Regularization of Convolutional Neural Networks with
  Cutout}.

\bibitem[Dosovitskiy et~al., 2014]{Dosovitskiy2014}
Dosovitskiy, A., Fischer, P., Springenberg, J.~T., Riedmiller, M., and Brox, T.
  (2014).
\newblock {Discriminative Unsupervised Feature Learning with Exemplar
  Convolutional Neural Networks}.

\bibitem[Gatys et~al., 2015]{Gatys2015}
Gatys, L.~A., Ecker, A.~S., and Bethge, M. (2015).
\newblock {A Neural Algorithm of Artistic Style}.

\bibitem[Geirhos et~al., 2018]{Geirhos2018}
Geirhos, R., Rubisch, P., Michaelis, C., Bethge, M., Wichmann, F.~A., and
  Brendel, W. (2018).
\newblock {ImageNet-trained CNNs are biased towards texture; increasing shape
  bias improves accuracy and robustness}.

\bibitem[Georgievski, 2019]{Georgievski2019}
Georgievski, B. (2019).
\newblock {Image Augmentation with Neural Style Transfer}.
\newblock pages 212--224.

\bibitem[Ghiasi et~al., 2017]{Ghiasi2017}
Ghiasi, G., Lee, H., Kudlur, M., Dumoulin, V., and Shlens, J. (2017).
\newblock {Exploring the structure of a real-time, arbitrary neural artistic
  stylization network}.

\bibitem[Gilpin et~al., 2018]{gilpin2018explaining}
Gilpin, L.~H., Bau, D., Yuan, B.~Z., Bajwa, A., Specter, M., and Kagal, L.
  (2018).
\newblock Explaining explanations: An overview of interpretability of machine
  learning.

\bibitem[Gkitsas et~al., 2019]{Gkitsas2019}
Gkitsas, V., Karakottas, A., Zioulis, N., Zarpalas, D., and Daras, P. (2019).
\newblock {Restyling Data: Application to Unsupervised Domain Adaptation}.

\bibitem[Hesse et~al., 2019]{Hesse2019a}
Hesse, L.~S., Kuling, G., Veta, M., and Martel, A.~L. (2019).
\newblock {Intensity augmentation for domain transfer of whole breast
  segmentation in MRI}.

\bibitem[Jackson et~al., 2018]{Jackson2018}
Jackson, P.~T., Atapour-Abarghouei, A., Bonner, S., Breckon, T., and Obara, B.
  (2018).
\newblock {Style Augmentation: Data Augmentation via Style Randomization}.

\bibitem[Ji et~al., 2018]{Ji2018}
Ji, X., Henriques, J.~F., and Vedaldi, A. (2018).
\newblock {Invariant Information Clustering for Unsupervised Image
  Classification and Segmentation}.

\bibitem[Jing et~al., 2017]{Jing2017}
Jing, Y., Yang, Y., Feng, Z., Ye, J., Yu, Y., and Song, M. (2017).
\newblock {Neural Style Transfer: A Review}.

\bibitem[Kabir et~al., 2020]{spinal2022net}
Kabir, H. M.~D., Abdar, M., Jalali, S. M.~J., Khosravi, A., Atiya, A.~F.,
  Nahavandi, S., and Srinivasan, D. (2020).
\newblock Spinalnet: Deep neural network with gradual input.

\bibitem[{Kotovenko, Dmytro, adn Sanakoyeu, Artsiom, and Lang, Sabine, and
  Ommer}, 2019]{kotovenko2019iccv}
{Kotovenko, Dmytro, adn Sanakoyeu, Artsiom, and Lang, Sabine, and Ommer}, B.
  (2019).
\newblock {Content and Style Disentanglement for Artistic Style Transfer}.

\bibitem[Li et~al., 2018]{Li2018}
Li, X., Liu, S., Kautz, J., and Yang, M.-H. (2018).
\newblock {Learning Linear Transformations for Fast Arbitrary Style Transfer}.

\bibitem[Selvaraju et~al., 2017]{selvaraju2017grad}
Selvaraju, R.~R., Cogswell, M., Das, A., Vedantam, R., Parikh, D., and Batra,
  D. (2017).
\newblock {Grad-cam: Visual explanations from deep networks via gradient-based
  localization}.
\newblock In {\em Proceedings of the IEEE International Conference on Computer
  Vision}, pages 618--626.

\bibitem[Shrikumar et~al., 2017]{shrikumar2017learning}
Shrikumar, A., Greenside, P., and Kundaje, A. (2017).
\newblock {Learning important features through propagating activation
  differences}.
\newblock In {\em Proceedings of the 34th International Conference on Machine
  Learning-Volume 70}, pages 3145--3153. JMLR. org.

\bibitem[Simonyan et~al., 2013]{simonyan2013deep}
Simonyan, K., Vedaldi, A., and Zisserman, A. (2013).
\newblock Deep inside convolutional networks: Visualising image classification
  models and saliency maps.

\bibitem[Sundararajan et~al., 2017]{sundararajan2017axiomatic}
Sundararajan, M., Taly, A., and Yan, Q. (2017).
\newblock {Axiomatic attribution for deep networks}.
\newblock In {\em Proceedings of the 34th International Conference on Machine
  Learning-Volume 70}, pages 3319--3328. JMLR. org.

\bibitem[Szegedy et~al., 2016]{Szegedy2016}
Szegedy, C., Ioffe, S., Vanhoucke, V., and Alemi, A. (2016).
\newblock {Inception-v4, Inception-ResNet and the Impact of Residual
  Connections on Learning}.

\bibitem[Szegedy et~al., 2015]{Szegedy2015}
Szegedy, C., Vanhoucke, V., Ioffe, S., Shlens, J., and Wojna, Z. (2015).
\newblock {Rethinking the Inception Architecture for Computer Vision}.

\bibitem[Tanaka and Aranha, 2019]{Tanaka2019}
Tanaka, F. H. K. d.~S. and Aranha, C. (2019).
\newblock {Data Augmentation Using GANs}.

\bibitem[Thoma, 2017]{Thoma2017}
Thoma, M. (2017).
\newblock {Analysis and Optimization of Convolutional Neural Network
  Architectures}.

\bibitem[Ulyanov et~al., 2017]{Ulyanov2017}
Ulyanov, D., Vedaldi, A., and Lempitsky, V. (2017).
\newblock {Improved Texture Networks: Maximizing Quality and Diversity in
  Feed-forward Stylization and Texture Synthesis}.

\bibitem[Zagoruyko and Komodakis, 2016]{Zagoruyko2016_1}
Zagoruyko, S. and Komodakis, N. (2016).
\newblock {Wide Residual Networks}.

\bibitem[Zeiler and Fergus, 2014]{zeiler2014visualizing}
Zeiler, M.~D. and Fergus, R. (2014).
\newblock {Visualizing and understanding convolutional networks}.
\newblock In {\em European conference on computer vision}, pages 818--833.
  Springer.

\bibitem[Zhao et~al., 2015]{Zhao2015}
Zhao, J., Mathieu, M., Goroshin, R., and LeCun, Y. (2015).
\newblock {Stacked What-Where Auto-encoders}.

\bibitem[Zheng et~al., 2019]{Zheng2019}
Zheng, X., Chalasani, T., Ghosal, K., Lutz, S., and Smolic, A. (2019).
\newblock {STaDA: Style Transfer as Data Augmentation}.

\bibitem[Zhou et~al., 2016]{zhou2016learning}
Zhou, B., Khosla, A., Lapedriza, A., Oliva, A., and Torralba, A. (2016).
\newblock {Learning deep features for discriminative localization}.
\newblock In {\em Proceedings of the IEEE conference on computer vision and
  pattern recognition}, pages 2921--2929.

\end{thebibliography}

}



\end{document}